\documentclass[pdflatex,sn-nature]{sn-jnl}

\usepackage{graphicx}%
\usepackage{multirow}%
\usepackage{amsmath,amssymb,amsfonts}%
\usepackage{amsthm}%
\usepackage{mathrsfs}%
\usepackage[title]{appendix}%
\usepackage{xcolor}%
\usepackage{textcomp}%
\usepackage{manyfoot}%
\usepackage{booktabs}%
\usepackage{algorithm}%
\usepackage{algorithmicx}%
\usepackage{algpseudocode}%
\usepackage{listings}%
\usepackage{float}%
\usepackage{tabularx, booktabs}

\theoremstyle{thmstyleone}%
\theoremstyle{thmstyletwo}%

\theoremstyle{thmstylethree}%

\begin{document}

\title[Article Title]{Computational Lesions in Multilingual Language Models Separate Shared and Language-specific Brain Alignment}


\author[1]{\fnm{Yang} \sur{Cui}}\email{yang.cui@manchester.ac.uk}
\equalcont{These authors contributed equally to this work.}

\author[1]{\fnm{Jingyuan} \sur{Sun}}\email{jingyuan.sun@manchester.ac.uk}
\equalcont{These authors contributed equally to this work.}

\author[1]{\fnm{Yizheng} \sur{Sun}}\email{yizheng.sun@manchester.ac.uk}

\author[1]{\fnm{Yifan} \sur{Wang}}\email{yifan.wang@manchester.ac.uk}

\author[4,5]{\fnm{Yunhao} \sur{Zhang}}\email{zhangyunhao2021@ia.ac.cn}

\author[2]{\fnm{Jixing} \sur{Li}}\email{jixingli@cityu.edu.hk}

\author[3]{\fnm{Shaonan} \sur{Wang}}\email{shaonan.wang@polyu.edu.hk}

\author[1]{\fnm{Hongpeng} \sur{Zhou}}\email{hongpeng.zhou@manchester.ac.uk}

\author[6]{\fnm{John} \sur{Hale}}\email{jthale@jhu.edu}

\author[4,5]{\fnm{Chengqing} \sur{Zong}}\email{cqzong@nlpr.ia.ac.cn}

\author*[1]{\fnm{Goran} \sur{Nenadic}}\email{gnenadic@manchester.ac.uk}

\affil[1]{\orgdiv{Department of Computer Science},
\orgname{The University of Manchester},
\orgaddress{\city{Manchester}, \country{UK}}}

\affil[2]{\orgdiv{Department of Linguistics and Translation},
\orgname{City University of Hong Kong},
\orgaddress{\city{Hong Kong}, \country{China}}}

\affil[3]{\orgdiv{Department of Language Science and Technology},
\orgname{The Hong Kong Polytechnic University},
\orgaddress{\city{Hong Kong}, \country{China}}}

\affil[4]{\orgdiv{State Key Laboratory of Multimodal Artificial Intelligence Systems},
\orgname{Institute of Automation, CAS},
\orgaddress{\city{Beijing}, \country{China}}}

\affil[5]{\orgdiv{School of Artificial Intelligence},
\orgname{University of Chinese Academy of Sciences},
\orgaddress{\city{Beijing}, \country{China}}}

\affil[6]{\orgdiv{Cognitive Science Department},
\orgname{Johns Hopkins University},
\orgaddress{\city{Baltimore}, \state{MD}, \country{USA}}}


\abstract{

How the brain supports language across different languages is a basic question in neuroscience and a useful test for multilingual artificial intelligence. Neuroimaging has identified language-responsive brain regions across languages, but it cannot by itself show whether the underlying processing is shared or language-specific. Here we use six multilingual large language models (LLMs) as controllable systems and create targeted ``computational lesions'' by zeroing small parameter sets that are important across languages or especially important for one language. We then compare intact and lesioned models in predicting functional magnetic resonance imaging (fMRI) responses during \textasciitilde100 minutes of naturalistic story listening in native English, Chinese and French (112 participants). Lesioning a compact shared core reduces whole-brain encoding correlation by 60.32\% relative to intact models, whereas language-specific lesions preserve cross-language separation in embedding space but selectively weaken brain predictivity for the matched native language. These results support a shared backbone with embedded specializations and provide a causal framework for studying multilingual brain-model alignment.


}


\maketitle

\section{Introduction}\label{sec1}

Understanding how language processing generalizes across languages matters for both neuroscience and artificial intelligence. The human brain can support comprehension in many languages, from the analytic syntax of Mandarin to the fusional morphology of French, even though these languages place different demands on learning and processing \cite{bornkessel2016importance, wei2023native}. This diversity raises a basic question: how can one biological system handle many distinct communication systems? Does the brain rely mainly on a common set of neural operations across languages, or does it also develop mechanisms tuned to the properties of a person’s native language?

Neuroimaging work has made important progress by identifying a fronto-temporo-parietal ``language network'' that is engaged across many languages \cite{malik2022investigation,fedorenko2024language,fedorenko2014reworking,hu2023precision}. This points to a shared anatomical scaffold for language. But shared anatomy does not by itself reveal shared processing. The same regions can be active while supporting different operations over different grammatical and lexical structures \cite{fedorenko2012lexical,regev2024high}. It therefore remains unclear whether the mechanisms implemented on this scaffold are largely common across languages or become partly specialized \cite{fedorenko2020broca}. This question is difficult to settle with traditional neuroimaging alone \cite{logothetis2008we}.

Large Language Models (LLMs) offer a complementary way to study this problem because they perform explicit computations that can be inspected and perturbed. Recent studies have begun to compare language models with brain recordings by asking whether representations extracted from a model can predict measured neural responses during language comprehension. These studies have shown that transformer-based language models can predict a substantial portion of neural responses during naturalistic language comprehension, suggesting that their internal representations capture features relevant to human language processing rather than merely task-specific outputs \cite{caucheteux2022deep, antonello2023scaling, schrimpf2021neural, kumar2024shared}. When representations from a model can predict recorded brain responses, this suggests that the model and the brain capture some aspects of language in similar ways; here we refer to this correspondence as \emph{brain-model alignment}. However, existing work has two important limitations. First, most studies rely on monolingual, English-based models \cite{schrimpf2021neural, joshi2020state}, which makes it difficult to determine whether the observed alignment reflects general language processing or properties specific to a single language. Second, although recent work has begun to manipulate or ablate parts of model representations \cite{schrimpf2021neural, jain2018incorporating, toneva2019interpreting, caucheteux2022deep}, these interventions have largely been developed in monolingual settings. As a result, it remains unclear whether the model components that support brain-model alignment are shared across languages or selective to individual languages.

Multilingual LLMs provide a direct testbed for these questions. Trained on large corpora spanning hundreds of languages, these models face a problem that mirrors the one faced by the brain: representing diverse linguistic systems within a single set of parameters \cite{goldstein2022shared, malik2022investigation, conneau2020unsupervised}. A single multilingual model can perform well across many language understanding tasks in many of its training languages, implying representations that generalize across languages \cite{conneau2020unsupervised,gaschi2023exploring}. At the same time, multilingual training could be implemented through shared processing, through language-specific subsystems, or through a mixture of both \cite{yang2019task}. This makes multilingual LLMs a controlled way to test whether multilingual language processing is supported by a shared core, by language-specific specializations, or by a mixture of both, and to compare these alternatives with competing hypotheses about how the brain supports multiple languages \cite{schrimpf2021neural, kriegeskorte2015deep}.

Here we introduce a causal test of brain-model alignment using computational lesions in multilingual language models from the LLaMa, Mistral and QWEN families \cite{touvron2023llama,bai2025qwen2,team2024qwen2,adler2024nemotron,muralidharan2024compact,sreenivas2024llm}. In this context, a computational lesion refers to a targeted disruption of selected model components, implemented by ablating specific parameter subsets \cite{zhang2024unveiling}. Most prior work establishes correspondence using intact models. We instead use targeted perturbations to ask which model components are necessary for that correspondence, and whether those components are shared across languages or selective to individual languages. To do this, we compare intact and lesioned multilingual models and examine how these perturbations change representational structure and voxel-wise fMRI predictivity during native-language story comprehension in English, Chinese and French. This framework is designed to distinguish among three possibilities: multilingual brain-model alignment may depend mainly on a shared core, mainly on language-specific components, or on a mixture of both. The results support the third possibility.

\section{Results}\label{sec2}

We first tested whether shared model components are necessary for brain predictivity across languages. We compared encoding performance between intact models and models carrying targeted core-language lesions during native-language story comprehension.

\begin{figure}[H]
    \centering
        
    \includegraphics[width=0.8\linewidth]{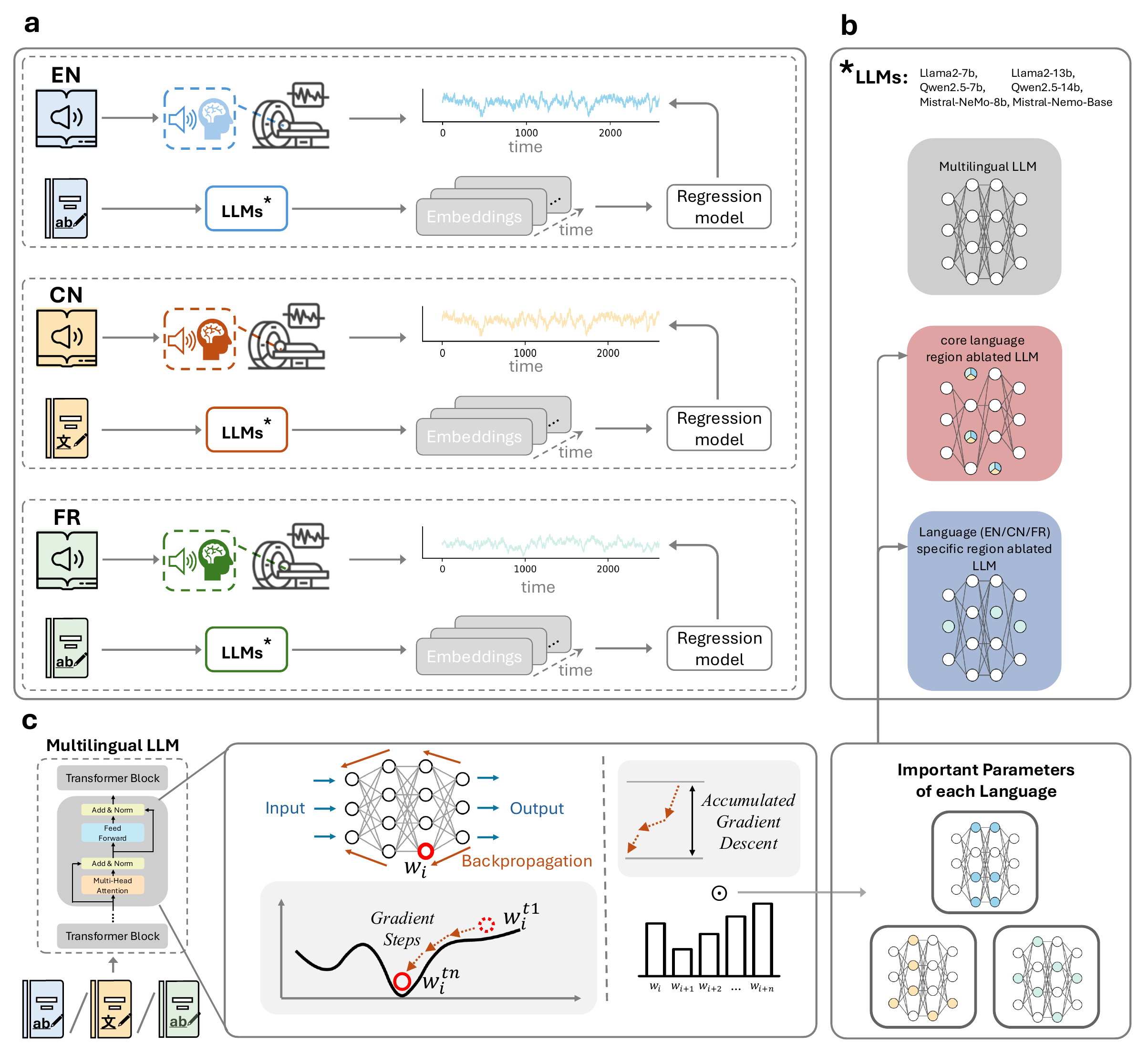}
    \caption{\footnotesize
    \textbf{Overview of the multilingual neural encoding framework with computationally lesioned large language models}\\ \textbf{(a), Multilingual neural encoding pipeline.} Native speakers of English (EN), Chinese (CN), and French (FR) listened to audiobook narratives while undergoing fMRI scanning. Textual stimuli were processed by large language models (LLMs) to extract contextualized embeddings from the final hidden layer, which were then used in voxel-wise encoding models to account for neural responses during native-language comprehension. \textbf{(b), Multilingual LLMs and lesion conditions.} For each model family, we evaluated an intact multilingual LLM together with computationally lesioned variants, including a core-language–ablated model and language-specific–ablated models corresponding to English, Chinese, and French. These models were used to dissociate shared versus language-specific contributions to neural encoding. \textbf{(c), Gradient-guided computational lesioning.} Core and language-specific parameter subsets within each multilingual LLM were identified using gradient-based importance estimates derived from language-specific full-parameter fine-tuning. During backpropagation, parameter gradients were accumulated and combined with their corresponding weights (gradient $\times$ weight) to quantify functional importance. Parameters ranking in the top 1\% of importance within each subset were then selectively ablated (zeroed) to induce targeted computational lesions, yielding core-lesioned and language-specific-lesioned models for downstream neural encoding analyses.
    }
    \label{fig:pipline}
\end{figure}

\subsection{Multilingual LLMs predict cortical responses across languages and model families}

We evaluated this framework on a multilingual naturalistic fMRI dataset in which native speakers of English, Chinese, and French listened to audiobook narratives from \textit{The Little Prince} in their respective native languages \cite{li2022petit}. For each participant, we fit voxel-wise encoding models over the cortical surface using final-layer embeddings from each multilingual LLM. The voxel-wise encoding models were fit using ridge regression with cross-validation, and performance was quantified as the Pearson correlation between predicted and observed BOLD time series on held-out data (Fig. \ref{fig:pipline}a) \cite{huth2016natural,wehbe2014simultaneously}.

To test robustness across architectures, we ran all analyses on six multilingual LLMs from three model families---LLaMA2, Qwen2.5, and Mistral---each at two parameter scales \cite{touvron2023llama,bai2025qwen2,team2024qwen2,adler2024nemotron,muralidharan2024compact,sreenivas2024llm}. For each model-language pair, we performed a voxel-wise one-sample t-test on the Pearson correlations across participants to identify significant voxels. We retained voxels passing a False Discovery Rate (FDR) corrected threshold of p $<$ 0.01 \cite{benjamini2001control}.

\begin{figure}[H]
    \centering
    \includegraphics[width=0.8\linewidth]{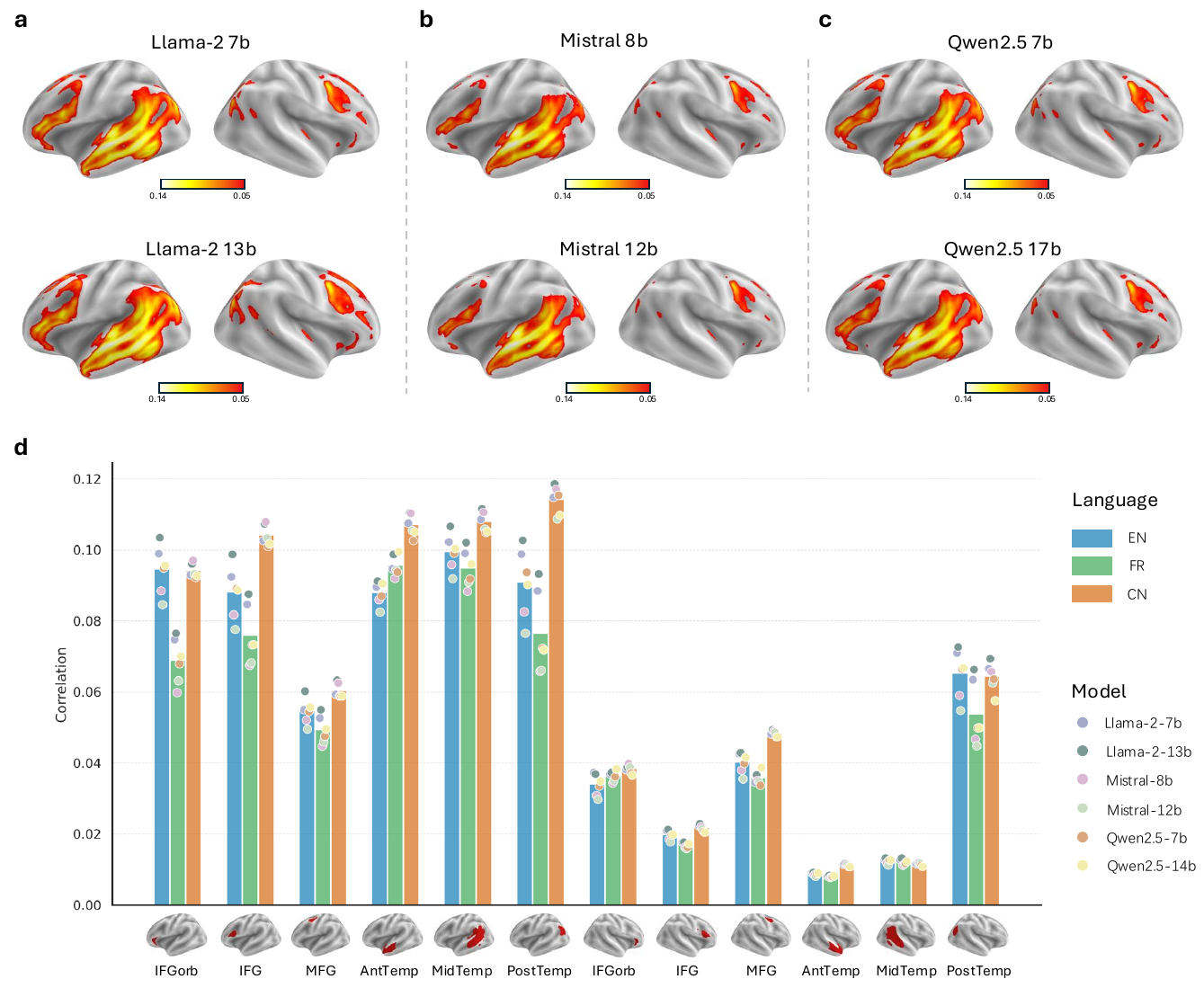}
    \caption{\footnotesize
    \textbf{Neural encoding accuracy across multilingual large language models}\\
    \textbf{a–c, Cross-linguistically robust cortical encoding maps across multilingual language models.} Cortical maps of voxel-wise encoding correlation for six large language models across English (EN), Chinese (CN), and French (FR) native listeners. For each model, we first computed voxel-wise neural encoding accuracy as the Pearson correlation between predicted and observed BOLD time series, averaged across all participants within each language. Voxels exceeding a significance threshold of $p<0.01$ were retained, and the intersection of significant voxels across the three languages was computed. The resulting shared voxel mask was then averaged across languages to generate a cross-linguistically robust cortical encoding map for each model.
    \textbf{d, Region-of-interest neural encoding performance across languages and models.} Model-wise neural encoding performance within functionally defined language regions of interest. Using the language network parcellation defined by Fedorenko and colleagues, each scatter point represents the mean encoding correlation of a single model within a given region and language, while bar plots indicate the across-model average performance. Across inferior frontal, middle frontal, anterior temporal, mid-temporal, and posterior temporal ROIs, all six models exhibited reliable predictive power, with broadly comparable performance across English, French, and Chinese. These results demonstrate that despite architectural and training differences, multilingual LLMs capture cortical representations that generalize across languages and align with established human language networks.
    }
    \label{fig:encoding_performance}
\end{figure}

We then intersected the significant voxel masks from English, Chinese, and French to obtain a cross-linguistic conjunction mask, and averaged encoding correlations within this shared cortical space to generate a robust cross-linguistic cortical encoding map for each model (Fig.~\ref{fig:encoding_performance}a-c).

Across model families, encoding performance was concentrated within the canonical language network, with strong predictivity in left-lateralized temporal regions, especially mid and posterior temporal cortex \cite{malik2022investigation,fedorenko2024language,hu2023precision}. Within the range of model scales examined here (approximately 7B to 13--14B parameters), increasing parameter count did not yield a systematic improvement in neural predictivity, suggesting that the observed brain-model alignment is not simply a function of model size.
We also summarized model performance in functionally defined language regions based on the parcellation of Fedorenko and colleagues \cite{fedorenko2024language} (Fig.~\ref{fig:encoding_performance}d). Overall, multilingual LLM embeddings reliably predicted neural responses within the core language network across all three languages. Within this network, two additional patterns stood out. First, encoding performance was generally higher in the left hemisphere than in the right, consistent with the well-known left-lateralization of language processing. There are previous works showing that high-level linguistic computations—such as syntactic composition, lexical access, and sentence-level semantic integration—are primarily supported by a distributed fronto-temporal network in the left hemisphere \cite{fedorenko2011functional,chai2016functional,ozernov2025precision}. In contrast, homologous regions in the right hemisphere tend to be less selective for core linguistic computations and are more strongly involved in broader contextual or social aspects of communication, such as discourse-level interpretation or pragmatic inference \cite{rajimehr2022complementary}. Because the internal representations of large language models primarily capture lexical and syntactic structure of text, their embeddings are expected to align more strongly with the computations carried out in the left-hemisphere language network, leading to higher encoding performance in these regions. Second, French showed weaker alignment in IFGorb, IFG, and posterior temporal cortex compared with English and Chinese, indicating cross-linguistic variability in how model representations map onto parts of the language network. Overall, these baseline results establish that multilingual LLM embeddings provide reliable predictors of cortical responses across languages and across model families, creating a stable starting point for causal tests with computational lesions.

\subsection{Core lesions collapse representational geometry and vanish shared brain alignment}
Using gradient information accumulated during full-parameter fine-tuning, we identified two complementary parameter subsets in each model: a shared core that was consistently important across languages, and language-specific subsets that were selectively important for individual languages \cite{zhang2024unveiling} (Fig.~\ref{fig:pipline}). To test the contribution of the shared core, we ablated the top 1\% of parameters within this cross-language subset. This manipulation produced a marked disruption of language-modeling ability. In Qwen2.5-7B, perplexity on the held-out Salesforce/Wikitext-2 test set increased from 6.75 in the intact model to 3,792, 672.75 after core ablation \cite{merity2016pointer,chen1999empirical,chelba2013one}. These results indicate that the identified core subset supports essential computations for multilingual language processing and motivate testing whether it is also necessary for brain alignment.

\begin{figure}[H]
    \centering
    \includegraphics[width=0.9\linewidth]{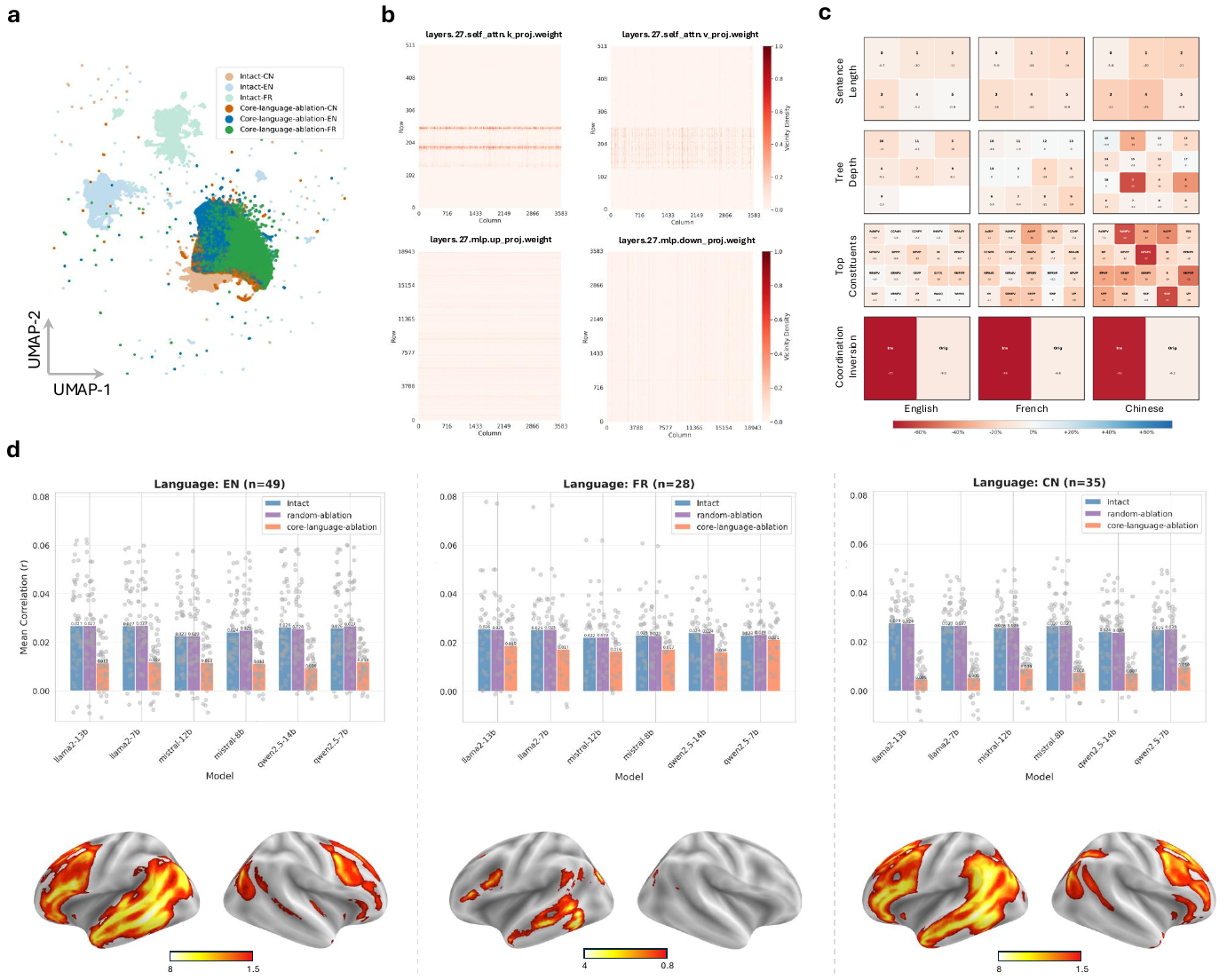}
    \caption{\footnotesize\textbf{Disrupting core language regions in multilingual LLMs alters internal representations and reduces neural predictivity.} \\
    \textbf{(a), Cross-linguistic representational shift after core language lesion.} UMAP visualizations of final-layer token embeddings from Qwen2.5–7B show distinct representational clusters for Chinese, English, and French in the intact model (light colors). After ablating the core language region (uniform 1\% parameter removal across layers and components), the embeddings of all three languages become compressed and partially overlapping (solid colors), indicating a systematic loss of representational structure. \textbf{(b), Parameter localization of the core language lesion.} Heatmaps illustrate the disrupted parameters within the final transformer layer (layer 27), highlighting the distribution of ablated weights across both self-attention (key and value projections) and MLP (up- and down-projection) components. \textbf{(c), Impaired linguistic competence following core language disruption.} Across four probing tasks (sentence length, tree depth, top constituents, and coordination inversion), performance decreases consistently in all three languages, demonstrating that the ablation selectively disrupts features closely tied to syntactic and hierarchical linguistic structure. \textbf{(d), Neural encoding consequences of core language lesion across languages.} Voxel-wise neural encoding analyses reveal substantial decreases in predictive performance when using embeddings from core-language–ablated LLMs. Bar plots show cross-participant mean encoding correlations for intact models (blue), random ablated model (purple) and core-language-region ablated models (orange) across six LLMs, with grey dots representing individual participants. Across English, French, and Chinese, intact models consistently outperform their ablated counterparts. Below each bar chart, whole-brain t-maps visualize voxel-wise differences between intact and ablated model predictions, with larger t-values indicating stronger reductions in neural predictivity following ablation. Reduced encoding performance is most prominent within bilateral temporal cortices and inferior frontal regions, demonstrating that the ablated parameters correspond to neural computations that are critical for supporting native language processing across languages.
    }
    \label{fig:core-language-ablation}
\end{figure}

We next examined how core ablation changes the geometry of final-layer embeddings. To visualize the organization of high-dimensional embeddings, we used Uniform Manifold Approximation and Projection (UMAP), a nonlinear dimensionality-reduction method that preserves local neighborhood structure in a low-dimensional space. In Qwen2.5-7B, UMAP projections show clear language-separated clusters in the intact model, but substantial compression and overlap after core ablation (Fig.~\ref{fig:core-language-ablation}a). This pattern was also evident in the original embedding space: the intact model showed a Silhouette Coefficient of 0.4277, whereas the core-language-ablated model showed 0.0594, a decrease of $\Delta = 0.3683$. Together, these results indicate that core ablation removes structure that helps maintain distinct language manifolds.

Core parameters were selected proportionally across all transformer layers and computational components, including self-attention projections (K/V) and MLP up/down projections, with 1\% removed per component per layer. This sampling makes the manipulation distributed rather than concentrated in a single layer or module \cite{tenney2019bert}. Fig.~\ref{fig:core-language-ablation}b shows the resulting distribution in the final layer of Qwen2.5-7B: within the attention block, ablated weights cluster in a subset of heads, while MLP ablations are more diffuse. This contrast is consistent with prior work linking attention and MLP components to different roles in transformer computation, without requiring that the effect be localized to a single component \cite{vaswani2017attention, geva2021transformer, elhage2021mathematical}.

We then tested whether core ablation disrupts linguistic information that is commonly tied to structure. Using multilingual probing datasets aligned across English, French, and Chinese, we evaluated sentence length, syntactic tree depth, top constituents, and coordination inversion (Fig.~\ref{fig:core-language-ablation}c). Sentence embeddings were computed by averaging token embeddings from the final layer, and MLP classifiers were trained following established probing methodology \cite{conneau2018senteval}. Performance dropped across all languages and tasks after ablation, with the largest drops in top constituent identification and coordination inversion, which index structural and compositional properties. These results indicate that the ablated parameters support more than surface-level lexical statistics.

Finally, we asked how core ablation affects brain-model alignment. Across nearly all models, ablating $\sim$1\% of core parameters sharply reduced neural encoding performance (Fig.~\ref{fig:core-language-ablation}d). The effect was strongest for English and Chinese and was weaker but consistent for French. Voxel-wise t-maps showed that reductions were concentrated in canonical language-selective territories, including anterior and posterior temporal cortex and inferior frontal cortex. Together, these results show that a compact core parameter subset is necessary for maintaining cross-language embedding structure and for sustaining robust brain predictivity \cite{frankle2018lottery, schrimpf2021neural}. The convergence of representational collapse, probing declines, and reduced neural encoding suggests that core computations in multilingual LLMs capture features that are also important for predicting cortical responses during native-language comprehension \cite{tang2024language, malik2022investigation, tuckute2024driving}.

\subsection{Language-specific lesions selectively impair language-matched neural encoding and reveal graded cross-language similarity}

We next asked whether parameter subsets that were selectively important for individual languages make correspondingly selective contributions to representation and brain alignment. Using the same gradient-guided procedure as above, we identified language-specific parameter subsets for English, Chinese, and French in each model and ablated the top 1\% of parameters within each subset \cite{zhang2024unveiling}. Unlike core ablation, which caused a global disruption of multilingual language modeling, language-specific ablation produced more selective and graded impairments while preserving overall model function. This dissociation motivated us to test whether language-specific lesions alter representational geometry and brain alignment in a language-matched manner.

\begin{figure}[H]
    \centering
    \includegraphics[width=0.8\linewidth]{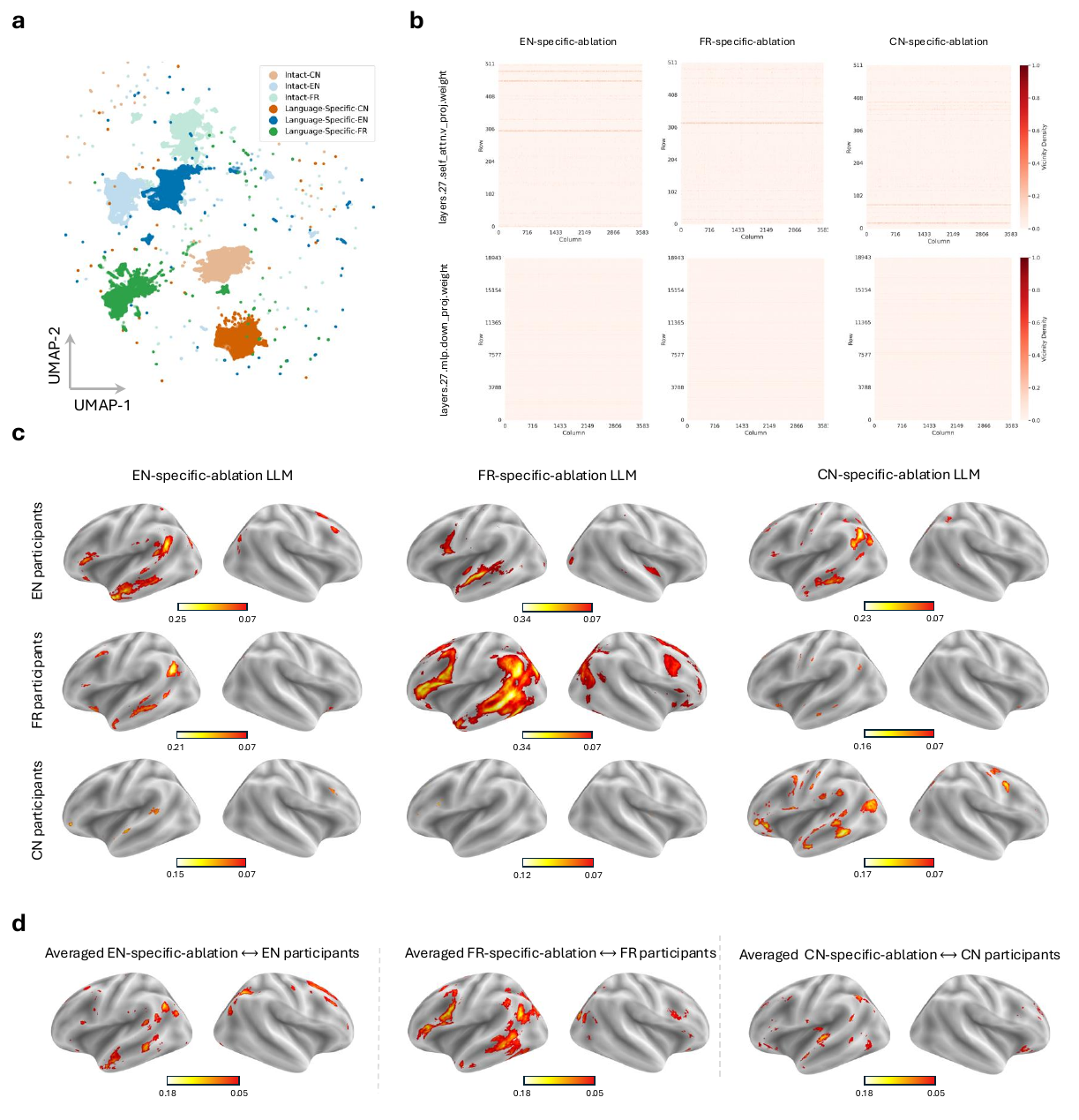}
    \caption{\footnotesize\textbf{Language-specific lesions selectively disrupt language-matched cortical encoding while preserving cross-language representational structure.}\\
    \textbf{a, Language-specific representational separation is preserved but reorganized after lesioning.}
    UMAP visualizations of final-layer token embeddings across English, French, and Chinese show well-separated clusters in the intact model (light colors). Following lesioning of each language-specific parameter subset, embeddings from the corresponding language (solid colors) shift to reorganized manifolds without collapsing onto other languages. This indicates that language-specific circuits support within-language representational structure rather than global cross-language separability.
    \textbf{b, Distribution of language-specific parameters across transformer components.}
    Heatmaps from the final transformer layer (layer 27) illustrate the spatial distribution of lesioned parameters for English-, French-, and Chinese-specific models. Language-specific parameters appear concentrated in structured rows within self-attention projections, whereas MLP components exhibit more spatially diffuse but systematic patterns, consistent with complementary roles in information routing versus representational transformation.
    \textbf{c, Language-matched neural encoding disruption following language-specific lesioning.}
    Voxel-wise paired t-tests comparing intact and lesioned models (Mistral-8B) were projected onto the cortical surface. Lesioning English-specific parameters selectively reduced neural predictivity in English participants, with analogous matched-language effects observed for French and Chinese. Disruptions localize to canonical language-selective regions, including anterior and posterior temporal cortex and inferior frontal gyrus. 
    \textbf{d, Convergent encoding impairment across models.}
    For each language, T-maps from language-matched lesions were averaged across all six models and summarized using a Language Processing Index. Group-level cortical maps reveal robust, left-lateralized reductions in neural predictivity, demonstrating that language-specific parameter subsets disproportionately support alignment with native-language brain responses.}
    \label{fig:language-specific-ablation}
\end{figure}

In Qwen2.5-7B, UMAP projections show that embeddings remain separated by language after language-specific lesions (Fig.~\ref{fig:language-specific-ablation}a). Instead of collapsing across languages, embeddings from the lesioned language shift within their own region of the space. This suggests that language-specific parameters contribute to within-language organization, whereas global separation across languages depends more strongly on shared core computations.

The parameter maps also differed across languages. In the final layer of Qwen2.5-7B (Fig.~\ref{fig:language-specific-ablation}b), English-, French-, and Chinese-specific ablations showed distinct spatial distributions across attention and MLP components. English- and French-specific parameters overlap more in the attention projection matrix, while Chinese-specific parameters occupy a largely non-overlapping range. This provides a structural view of language selectivity within the model and is consistent with partly distinct subcircuits across languages \cite{tang2024language}.

We then tested whether these language-specific components matter for brain alignment in a language-matched way. Using Mistral-8B as an example, voxel-wise t-tests comparing intact and lesioned models showed a clear diagonal pattern (Fig.~\ref{fig:language-specific-ablation}c): language-matched ablations produced larger reductions in encoding performance than mismatched ablations. This indicates that language-specific parameters in the model contribute most strongly to predicting cortical responses in native speakers of the same language.

These effects were smaller than those observed for core ablation, consistent with a dominant shared system alongside more limited language-specific contributions \cite{malik2022investigation}. We also observed asymmetries across language pairs. English- and French-specific ablations affected encoding in both English and French participants more strongly than in Chinese participants, while Chinese-specific ablation showed weaker cross-language effects. This pattern is consistent with graded similarity between languages, given that English and French share many properties within the Indo-European family, while Chinese is typologically more distant \cite{pires2019multilingual, kudugunta2019investigating}. In this view, language-specific representations need not form strictly isolated streams; instead, they can reflect partial overlap that tracks linguistic similarity \cite{chang2022geometry}.

To summarize where language-specific lesion effects manifest on the cortex, we computed a voxel-wise Language Processing Index (LPI) that contrasts neural encoding performance between intact models and their language-specific–lesioned counterparts. 
For each model and native-language participant group, voxel-wise paired $t$-tests were first conducted comparing encoding performance derived from intact and lesioned models, yielding three lesion-specific $t$-maps per architecture (English-, Chinese-, and French-targeted lesions). To ensure cross-language comparability, $t$-maps within each model were normalized using min–max scaling across cortical voxels. LPI values were then computed for each target language (see Eq.~\ref{eq:lpi}), quantifying the relative degradation of encoding alignment induced by lesioning that language compared with the remaining languages. Finally, voxel-wise LPI maps were averaged across the six models to obtain architecture-robust projections of language-specific effects in standard space.
The interpretational logic of the LPI is inherently relative: voxels receive high index values when lesioning the target language produces disproportionately stronger encoding disruption than lesioning other languages at the same location. This formulation isolates language-dependent cortical sensitivities while factoring out shared multilingual encoding substrates.
LPI maps (Fig.~\ref{fig:language-specific-ablation}d) localized language-specific effects primarily within the broader fronto–temporo–parietal language network, but in sparser and more spatially fragmented clusters compared with the distributed reductions observed under core-language ablation. After averaging across models, these effects emerged as discrete cortical patches rather than a contiguous network, suggesting that language-specific processing is implemented through localized functional specializations embedded within a largely shared cortical infrastructure \cite{benson1999language,caucheteux2021disentangling,lescroart2019human}.

\section{Discussion} \label{sec:discussion}

A central question in multilingual neuroscience is whether native-language comprehension is supported by largely shared neural computations across languages or whether it also recruits language-specific cortical mechanisms \cite{bolger2005cross, perani2005neural}. The present results suggest that both are true, but at different levels of organization. Across model families, intact multilingual LLMs provided reliable predictors of cortical responses in English, Chinese and French, establishing a robust baseline for multilingual brain--model alignment. Within this baseline, lesioning a compact shared core of model parameters caused a broad collapse of representational geometry and a marked reduction in neural predictivity across languages, whereas lesioning language-specific parameter subsets left global multilingual structure largely intact but selectively weakened alignment for the matched native language. These language-specific effects were smaller and more spatially restricted than the effects of core lesions, and they showed graded overlap across languages, with English and French patterns more similar to one another than to Chinese. Together, these findings support a shared computational backbone with embedded language-specific specializations.

Most model-to-brain studies have approached this question correlationally---for example, by comparing alignment across model layers or architectures \cite{schrimpf2021neural, caucheteux2022brains, jain2018incorporating}---which leaves open whether particular computational components are \emph{causally necessary} for the observed correspondences \cite{saxe2021if, bowers2023deep}. Here, we take a mechanistic route: by introducing gradient-guided computational lesions into multilingual LLMs, we test how targeted disruptions alter voxel-wise fMRI predictivity during naturalistic story comprehension. This shift in logic matters. Instead of treating brain predictivity as a single score to maximize, we use changes in predictivity under controlled disruptions to constrain what types of computations the cortical signal depends on.

\subsection{What lesion-based evidence adds to brain-model alignment}

Alignment between model embeddings and brain activity can arise for several reasons, including shared sensitivity to broad properties of the stimulus or the ability of a linear mapping to extract weak signal spread across features \cite{schrimpf2021neural, caucheteux2022brains}. Lesion-based analyses narrow this interpretive space by asking a different question: when a specific subset of parameters is disrupted, does predictivity fail in a systematic way, and does that failure localize to language-selective cortical territories? If so, then the disrupted subset is not merely correlated with the brain signal; it is part of what the mapping relies on.

This supports a precise, limited claim. The results do not imply that cortex implements the same algorithm as a transformer. Rather, they show that certain representational properties supported by specific parameter subsets in multilingual LLMs are also required to predict cortical responses during comprehension. In this sense, lesions move the discussion from ``which intact representations fit'' to ``which internal computations must remain intact for the fit to hold'' \cite{saxe2021if, bowers2023deep}. That is a stronger kind of constraint than layer-wise comparisons alone, because it ties the correspondence to a failure mode that can be reproduced across architectures.

\subsection{A sparsity principle for brain-aligned computation}

A major implication is that brain-model alignment is concentrated in the parameter space. Using full-parameter fine-tuning and gradient--weight importance estimates, we identify compact parameter subsets and ablate only the top 1\% within each subset (core and language-specific) \cite{zhang2024unveiling}. Despite being small, this intervention produces large, linked changes at three levels: internal representation geometry, linguistic competence, and neural encoding.

The clearest case is the core language region. Disrupting it yields a convergent ``triple collapse'': (i) cross-language representational structure compresses and overlaps (e.g., a sharp drop in Silhouette Coefficient after core ablation in Qwen2.5-7B), (ii) linguistic competence catastrophically degrades, 
and (iii) brain predictivity sharply declines, with a reported mean 60.32\% decrease in whole-brain encoding correlation relative to intact models. The important point is not only that performance drops, but that the drop is structured: representational collapse, behavioral degradation, and loss of neural predictivity co-occur under the same compact perturbation. This co-occurrence suggests that the representational features driving brain predictivity are not arbitrary by-products; they are tied to computations that are central for the model's ability to represent language at all.

This concentration also changes how it is useful to talk about ``where'' alignment lives. Layer-wise analyses implicitly suggest that alignment is a property of a layer or a block. The lesion results instead support a different description: alignment depends strongly on a relatively small subset of parameters whose influence propagates through the representation. This view naturally fits the fact that core parameters are distributed across layers and components: a small set of weights can still shape a computation that is expressed broadly in activations. In practical terms, it suggests that targeted perturbations can be used to generate sharper, testable hypotheses about which representational properties matter for cortical prediction.

\subsection{Shared core and language-specific circuits: a hybrid organization with graded similarity}

The results also argue against two extreme accounts. One is that native-language comprehension relies on the same computation in the same way for all languages; the other is that each language relies on separate, largely independent mechanisms. Instead, the findings support a hybrid organization.

On the model side, the importance mapping separates parameters that are consistently important across languages (the shared core) from those whose importance is selectively elevated for one language (language-specific regions). On the brain side, the consequences follow the same division. Core ablation produces broad reductions in predictivity across English, Chinese, and French, concentrated in canonical language-selective territories (bilateral temporal cortex and inferior frontal cortex) \cite{hu2023precision,mollica2020composition,malik2022investigation,fedorenko2012language}. In contrast, language-specific ablation leaves cross-language separability intact at the embedding level while selectively weakening brain predictivity most strongly when the ablation matches the participants' native language.

This pattern helps refine what anatomical overlap across languages can mean. Shared activation of the same macro-scale regions has long been compatible with both shared and distinct computations. The lesion results support a middle hypothesis: a dominant shared backbone that drives much of the mapping between model representations and the language network, alongside embedded specializations that modulate the mapping for particular language groups \cite{abutalebi2007bilingual,green2013language,fedorenko2011functional}. Importantly, this does not require language-specific processing to be isolated into separate, dedicated streams. The observed graded similarity---for example, stronger overlap between English and French than between either and Chinese---fits a picture in which ``language-specific'' effects reflect selective weighting within a largely shared system, with partial transfer among related languages rather than strict modular separation. This hybrid view also offers a natural explanation for why language-specific effects are consistently smaller than core effects: they are additions or adjustments on top of a shared computation, not replacements for it.

\subsection{Why distributed lesions matter for interpretation and localization}

A practical challenge in lesion-based arguments is to rule out trivial explanations such as ``damage to one critical layer'' or ``general disruption caused by removing weights.'' Two aspects of the design help here. First, core parameters are selected proportionally across layers and components, which makes the perturbation distributed and avoids concentrating damage in a single layer. Second, the random-ablation control (intact vs.\ random-ablated vs.\ core-ablated) separates effects of structured lesioning from effects of parameter removal per se.

These choices also matter for the brain-side interpretation. Lesions are most informative when they lead to spatially structured changes in encoding, rather than uniform degradation. The fact that encoding losses concentrate in language-selective territories supports the claim that the disrupted features are relevant to language comprehension as captured by these cortical regions, rather than reflecting indiscriminate loss of signal. The Language Processing Index (LPI) further sharpens localization by contrasting how much encoding degrades under language-specific ablation relative to core ablation, highlighting regions whose predictivity is disproportionately sensitive to language-specific disruptions \cite{caucheteux2021disentangling}.

\subsection{Robustness across models and implications of scale}

Another feature of the evidence is convergence across architectures. Analyses are replicated across six LLMs drawn from three families (LLaMA2, Qwen2.5, Mistral \cite{touvron2023llama,bai2025qwen2,team2024qwen2,adler2024nemotron,muralidharan2024compact,sreenivas2024llm}) and two parameter scales per family. This reduces the chance that the findings are an artifact of one tokenizer, one training distribution, or one architectural choice. Within the tested range ($\sim$7B to $\sim$13--14B), increased parameter count does not produce a systematic increase in neural predictivity. One interpretation is that, at least in this regime, brain predictivity depends more on having the right kinds of representational constraints than on simply increasing capacity. This strengthens the case for focusing on which computations support alignment, rather than treating scale as the primary explanation.

The neuroimaging setting also matters. The LPPC-fMRI dataset uses $\sim$100 minutes of naturalistic listening to \textit{The Little Prince} in each participant's native language (112 participants across EN/CN/FR) \cite{li2022petit}. Naturalistic stimuli increase ecological validity, but they also bring variability (e.g., translation choices, prosody, narrative timing). The lesion-based contrasts partly address this because they are computed within each language group, comparing intact and lesioned models on the same stimulus. This reduces reliance on cross-language comparability of the stimulus itself when making the main claims about shared versus language-specific dependencies.

\subsection{Limitations and alternative interpretations}

Several limitations are important for interpreting what the lesions do and do not establish.

\paragraph{Capability degradation as a confound (core ablation).}
Core ablation causes extreme perplexity inflation, so reduced neural predictivity could partly reflect global representational corruption rather than selective removal of brain-relevant computation \cite{schrimpf2021neural}. The random-ablation control and the spatial concentration of encoding losses in language-selective cortex reduce the plausibility of a purely nonspecific explanation, but they do not fully resolve it. A clean way to sharpen causal specificity is to add dose--response lesioning (varying ablation percentages) and matched-performance controls that hold overall language modeling degradation constant while varying which parameters are removed \cite{tuckute2024driving, ravfogel2020null}. These additions would help distinguish effects that track general competence loss from effects that reflect the removal of specific representational features needed for brain predictivity.

\paragraph{What ``core'' and ``language-specific'' mean operationally.}
The decomposition is defined by importance under language-specific fine-tuning objectives \cite{zhang2024unveiling}. This means that ``core'' should be read operationally as parameters that are consistently important across languages under the matched objectives, not as a guarantee of a single, language-agnostic algorithm. Likewise, ``language-specific'' denotes parameters with selective importance elevation, not necessarily parameters that are exclusively used by one language. The hybrid and graded patterns are consistent with this operational view: selective effects can arise from differences in weighting within shared computation as well as from more separated subcircuits.

\paragraph{Stimulus and language coverage.}
The design uses three languages and a single narrative, which supports direct comparisons \cite{honey2012not} but may entangle language differences with translation and speech properties \cite{hamilton2020revolution}. Extending to a wider range of typologies \cite{blasi2022over} and to controlled stimuli \cite{fedorenko2011functional} would test the generality of the hybrid organization and the graded similarity pattern.

\paragraph{Model-to-brain mapping choices.}
The encoding model is linear and uses final-layer embeddings, which may miss nonlinear correspondences \cite{ivanova2021simple} or layer-specific temporal dynamics \cite{toneva2019interpreting}. Future work could add layer-wise lesions and nonlinear encoders, and combine fMRI with temporally precise recordings to resolve when shared versus language-specific computations contribute \cite{brodbeck2018neural, gwilliams2020neural}. These extensions would not change the core logic of lesion-based inference, but they could sharpen which representational properties matter most, and when.

\paragraph{Scanner/site differences.}
English and Chinese data were collected on a GE system while French was collected on a Siemens system, and French shows weaker alignment in some regions in the ROI analyses. While lesion patterns remain consistent, harmonized acquisition or within-site multilingual cohorts would strengthen cross-language comparisons \cite{nastase2021narratives, yamashita2019harmonization}. The main lesion contrasts are computed within language groups, which reduces dependence on absolute comparability across sites, but does not eliminate it.

\subsection{Outlook}

The lesion results suggest a broader direction for brain-AI alignment work: treat multilingual LLMs as controllable computational systems, identify compact parameter circuits that matter for specific representational outcomes, and use lesion-induced changes to localize cortical dependencies. Several immediate extensions follow naturally. Testing bilinguals and L2 learners would help determine whether language-specific circuits track proficiency and exposure. Expanding language coverage would test whether graded similarity persists beyond English, French, and Chinese. Finer-grained lesions (e.g., attention heads vs.\ MLP blocks) could clarify which computational elements most strongly support brain predictivity. Together, these steps would help turn alignment from a descriptive observation into a set of falsifiable claims about which computations drive it.

\subsection{Conclusion}

Overall, the findings support an account in which native-language comprehension is dominated by a shared cortical computation that aligns with a compact, distributed core parameter circuit in multilingual LLMs, while additional language-specific circuits contribute selectively to alignment in native speakers of particular languages. The key conceptual move is not ``shared regions versus separate regions,'' but ``shared backbone with embedded specializations''---and the key methodological move is to establish this distinction using causal computational lesions rather than correlation alone.

\section{Methods}\label{sec11}
In this section, we describe our computational lesion paradigm, which leverages multilingual large language models (LLMs) to causally investigate shared and language-specific neural representations for language processing. Our approach consists of three main stages. First, we adopt the previous method and isolate a ``core'' multilingual parameter subnetwork and several language-specific subnetworks within LLMs by fine-tuning them on a next-word prediction task across three languages \cite{zhang2024unveiling,foroutan2022discovering}. Second, we systematically ablate these identified subnetworks to create ``lesioned'' models \cite{zhang2024unveiling}. Finally, we employ a neural encoding framework to quantify and compare the performance of intact versus lesioned models in predicting fMRI activity from native Chinese, English, and French speakers, thereby inferring the causal role of these subnetworks in aligning with human brain function and localizing their corresponding shared or specific cortical substrates \cite{dehghani2017decoding}.

\subsection{Comparison of multilingual LLM architectures used in this study}

To ensure robustness and generalizability, all analyses were conducted across six multilingual LLMs drawn from three model families—LLaMA2, Qwen2.5, and Mistral—each \cite{touvron2023llama,bai2025qwen2,team2024qwen2,adler2024nemotron,muralidharan2024compact,sreenivas2024llm} evaluated at two parameter scales. These model families were selected to reflect complementary linguistic emphases relevant to the trilingual encoding task \cite{xu2025survey}. LLaMA2, a widely adopted benchmark model developed by the USA company Meta, is predominantly trained on English-language data while maintaining multilingual capabilities \cite{touvron2023llama}. Qwen2.5, an open-source model pretrained with mainly English and Chinese corpus developed by Chinese company Alibaba, demonstrates strong performance on Chinese benchmarks and supports more than 29 languages, including French, reflecting substantial high-quality non-English training data\cite{bai2025qwen2,team2024qwen2}. Mistral models, developed by French company Mistral AI, place particular emphasis on multilingual proficiency, with reported native-level fluency across several European languages, including French \cite{adler2024nemotron,muralidharan2024compact,sreenivas2024llm}. Consistent neural encoding patterns observed across model families and sizes indicate that the identified shared and language-specific cortical representations are not driven by idiosyncratic properties of any single architecture or training distribution \cite{caucheteux2022brains, schrimpf2021neural, kornblith2019similarity, conneau2020emerging}.

\begin{table*}[t]
\centering
\label{tab:model_comparison}

\small
\setlength{\tabcolsep}{4pt}
\begin{tabularx}{\textwidth}{ll ccc c c c}
\toprule
\textbf{Model} & \textbf{Family} & \textbf{Params} & \textbf{Layers} & \textbf{Dim} & \textbf{Attention} & \textbf{Tokenizer} & \textbf{Ctx.} \\
\midrule

LLaMA2-7B   & LLaMA2  & 7B  & 32 & 4096 & MHA & SentencePiece & 4k   \\
LLaMA2-13B  & LLaMA2  & 13B & 40 & 5120 & MHA & SentencePiece & 4k   \\
Qwen2.5-7B  & Qwen2.5 & 7B  & 32 & 4096 & GQA & BPE           & 128k \\
Qwen2.5-14B & Qwen2.5 & 14B & 48 & 5120 & GQA & BPE           & 128k \\
Mistral-8B  & Mistral & 8B  & 40 & 4096 & GQA & SentencePiece & 8k   \\
Mistral-12B & Mistral & 12B & 40 & 5120 & GQA & SentencePiece & 8k   \\
\bottomrule
\end{tabularx}
\caption{\textbf{Architectural comparison of multilingual LLMs used in this study.} \\All models adopt decoder-only Transformer architectures but differ in scale(7-14B), attention design(MHA vs. GQA), tokenizer construction ..., and multilingual training emphasis. Architectural specifications are compiled from official technical reports and model documentation \cite{touvron2023llama,bai2025qwen2,team2024qwen2,adler2024nemotron,muralidharan2024compact,sreenivas2024llm}.}
\vspace{2pt}
\footnotesize\emph{Note.} Dim: Hidden Dimension; Ctx.: Context Length.
\end{table*}

All six LLMs used in this study adopt a transformer-based, decoder-only architecture, enabling direct comparability across model families. Owing to computational constraints associated with full-parameter fine-tuning, we restricted our analyses to two representative model scales—approximately 7B and 14B parameters—resulting in a total of six models across three families: LLaMA2, Qwen2.5, and Mistral.

Despite this shared backbone, the model families differ substantially in architectural configuration shown in Table \ref{tab:model_comparison}. LLaMA2 employs standard multi-head attention (MHA) with matched query and key–value heads, scaling performance primarily through depth and hidden width. Qwen2.5 adopts grouped-query attention (GQA) and integrates an expanded tokenizer and extended context window, optimizing multilingual density and long-range dependency modeling. Mistral models combine GQA with efficiency-oriented scaling, featuring large per-head dimensionality and multilingual tokenization optimized for European languages \cite{touvron2023llama,bai2025qwen2,team2024qwen2,adler2024nemotron,muralidharan2024compact,sreenivas2024llm}.

The convergence of neural encoding results across these heterogeneous architectures indicates that observed shared and language-specific cortical patterns are not driven by idiosyncratic design properties of any single model family.

\subsection{Identification and ablation of core and language-specific language representations}

To dissociate shared and language-specific linguistic representations within multilingual LLMs, we adopted a parameter-level ablation strategy \cite{bau2017network} grounded in established methods from model interpretability and theoretical neuroscience \cite{frankle2018lottery, sternberg2011modular}.

Figure \ref{fig:pipline}b-c has shown that for each dense multilingual LLM, we performed full-parameter fine-tuning separately on free-text corpora in English, Chinese, and French. During fine-tuning, we accumulated the gradients of each parameter across optimization steps. Following prior work in model interpretability \cite{molchanov2019importance,zhang2022platon}, we quantified the importance of a parameter for a given language as the product of its accumulated gradient magnitude and its original weight value. This quantity captures both the sensitivity of the loss to the parameter and the parameter’s contribution to the model’s computation, providing a principled estimate of functional relevance.

Using the resulting language-specific importance maps, we identified two complementary parameter sets. Parameters exhibiting consistently high importance across the three languages were defined as constituting a core language region, reflecting shared computational resources for multilingual language processing \cite{conneau2020unsupervised, kudugunta2019investigating, libovicky2020language}. Conversely, parameters whose importance was selectively elevated for a single language were designated as language-specific regions, corresponding to representations preferentially engaged during processing of that language \cite{singh2019bert, foroutan2022discovering, zhang2024unveiling}.

To causally test the functional role of these regions, we performed targeted ablation by uniformly disrupting the top 1\% of parameters within each identified region, separately for each transformer layer and architectural component. Ablation was implemented by setting the selected parameter values to zero \cite{morcos2018importance}, yielding one core language region–ablated model and three language-specific ablated models. This uniform, percentile-based procedure avoids confounding effects of uneven parameter density across layers \cite{gale2019state} and ensures comparability across models and scales \cite{blalock2020state}.

The effectiveness of this ablation strategy was validated using perplexity on next-word prediction. Core language region ablation resulted in a dramatic increase in perplexity—by several orders of magnitude relative to the intact model—indicating a severe degradation of linguistic competence and confirming that the ablated parameters are critical for language processing. In contrast, language-specific ablation produced more selective impairments, consistent with partial preservation of shared linguistic structure.

\subsubsection{Identification of Core and Language-Specific Subnetworks in LLMs} 
\paragraph{Models and Corpora} We selected three prominent families of multilingual LLMs: Llama 2, Qwen 2.5, and Mistral-Nemo \cite{touvron2023llama,bai2025qwen2,team2024qwen2,adler2024nemotron,muralidharan2024compact,sreenivas2024llm}. For each family, we utilized two model sizes to assess scalability: Llama2-7B and -13B, Qwen2.5-7B and -14B, and Mistral-Nemo-Minitron-8B and -Base-12B. These models represent state-of-the-art open-source architectures. All models are decoder-only Transformers, optimized for next-token prediction, but differ in architectural design and training strategy.
In all cases, we deliberately employed the base pretrained models rather than instruction-tuned or chat-oriented variants. This choice ensures that our neural encoding analyses target the core linguistic representations learned through unsupervised next-token prediction, without interference from alignment or reinforcement learning objectives. By comparing these model families, we aim to assess the generality and stability of language-to-brain mappings across architectures and linguistic typologies.


For the Chinese corpus, we curated canonical narrative works spanning historical, philosophical, and adventure genres, including Dream of the Red Chamber, Romance of the Three Kingdoms, and Twenty Thousand Leagues Under the Seas (Chinese translations). These texts provide dense character interactions, complex socio-temporal event chains, and descriptive passages conducive to eliciting structured linguistic representations.
For the English corpus, we selected 19th-century literary novels with comparable narrative continuity and psychological depth, namely Jane Eyre and Pride and Prejudice. Both works feature rich internal monologue, dialogue structure, and evolving interpersonal dynamics, which are valuable for modeling discourse-level representations.
For the French corpus, we compiled texts from L'Étranger and Les Misérables, capturing stylistic diversity from existentialist minimalism to expansive socio-historical narration. This combination enables the models to encounter varied syntactic constructions and narrative pacing within the same language.
All corpora were digitized, normalized, and segmented into continuous passages suitable for full-parameter fine-tuning.

\paragraph{Parameter Importance Estimation} We quantified the functional importance of each model parameter for each language through a full-parameter fine-tuning procedure. Each base LLM was independently fine-tuned on the Chinese, English, and French corpora using a standard next-word prediction objective with an autoregressive loss. During this process, we estimated the importance of each parameter $\theta_i$ by calculating the product of its magnitude and its accumulated absolute gradients \cite{molchanov2019importance,zhang2022platon}. Specifically, for a given language $L$, the importance score $I_L(\theta_i)$ is formally defined as: 

\begin{equation} I_L(\theta_i) = |\theta_i| \cdot \sum_{t} \left|\frac{\partial \mathcal{L}_L}{\partial \theta_i^{(t)}}\right| \label{eq:importance} 
\end{equation} 

where $\mathcal{L}_L$ is the loss function for language $L$ and $t$ indexes the training steps. This formulation captures the principle that a parameter is considered important if it both has a large weight magnitude ($|\theta_i|$) and is subject to significant and frequent updates during training ($\sum|\text{gradient}|$).

\paragraph{Core and Language-Specific Subnetworks} To identify the shared multilingual subnetwork, we defined a ``core'' importance score, $I_{\text{Core}}(\theta_i)$, by summing the parameter's importance scores across all three languages: 

\begin{equation} I_{\text{Core}}(\theta_i) = I_{\text{CN}}(\theta_i) + I_{\text{EN}}(\theta_i) + I_{\text{FR}}(\theta_i) \label{eq:core} 
\end{equation} 

We then ranked all parameters by this core importance score and designated the top 1\% as the ``Core Language Region''. To identify language-specific subnetworks, we computed a relative importance score, $I_{\text{Specific}, L}(\theta_i)$, which quantifies how much more important a parameter is for a target language compared to the others. This score is defined as:

\begin{gather*}
    P_{p,L} = \frac{Rank_{p,L} - 1}{N - 1} \\
    S^{rank}_{p,A} = P_{p,A} - \max(P_{p,B}, P_{p,C})
\end{gather*}

For each language, we ranked all parameters by their corresponding specific importance score and designated the top 1\% as the ``Language-Specific Region''. 
\paragraph{Computational Lesion Procedure} Based on the identified subnetworks, we created a series of lesioned models for each base LLM. The ``Core-Ablated LLM'' was generated by setting the weights of all parameters within the Core Language Region to zero \cite{morcos2018importance}. Similarly, three separate models—the ``Chinese-Specific Ablated LLM'', ``English-Specific Ablated LLM'', and ``French-Specific Ablated LLM''—were created by ablating the corresponding language-specific regions. 

\paragraph{Language Processing Index (LPI) and cross-model convergence}

To quantify the specificity of cortical responses for processing a given language relative to others, we defined a Language Processing Index (LPI) and evaluated its robustness across different large language model (LLM) architectures \cite{benson1999language}.

For each language (Chinese, English, and French) and each model, we first quantified context sensitivity by comparing neural encoding performance between the intact model and the corresponding language-specific ablation model \cite{toneva2019interpreting, jain2018incorporating}. Voxel-wise paired t-tests were performed across participants on encoding accuracy (Pearson correlation, r), yielding statistical parametric maps (t-maps) \cite{friston1994statistical} that capture the degree to which intact representations outperform ablated ones.

For each model, T-maps from the three languages were restricted to an MNI152 cortical mask. To account for systematic differences in T-value magnitude across languages and models, we applied Min–Max normalization to scale T-values to the [0, 1] range and rectified negative values to zero \cite{pereira2009machine, misaki2010comparison}. This normalization ensures that the LPI captures relative language specificity rather than absolute effect size differences \cite{fedorenko2011functional}.

For a given target language ($L_{target}$), the LPI was calculated voxel-wise as:
\begin{equation}
\label{eq:lpi}
LPI_v^{(L_{target})} =
\frac{
T_v^{(L_{target})} - \overline{T}_v^{(\mathrm{others})}
}{
T_v^{(L_{target})} + \overline{T}_v^{(\mathrm{others})} + \epsilon
}
\end{equation}
where $\bar{T}_v^{(\mathrm{others})}$ denotes the average normalized T-value of the other two languages and $\epsilon$ is a small constant added for numerical stability. Higher LPI values indicate greater specificity of a cortical region for processing the target language. To ensure interpretability, LPI maps were further masked by the significance map of the target language (p $<$ 0.01), retaining only voxels exhibiting reliable context sensitivity.

To identify language-specific cortical regions that generalize beyond any single model architecture, we computed voxel-wise averages of LPI maps across six LLMs (LLaMA2–7B/13B, Qwen2.5–7B/14B, and Mistral–8B/12B \cite{touvron2023llama,bai2025qwen2,team2024qwen2,adler2024nemotron,muralidharan2024compact,sreenivas2024llm}). The resulting cross-model averaged LPI maps reveal language-specific cortical patterns that are consistent across architectures and parameter scales, providing a robust estimate of language-selective neural representations.

\paragraph{Embedding geometry analysis}

To quantify how core lesions alter multilingual representational geometry (Figure \ref{fig:core-language-ablation}a, \ref{fig:language-specific-ablation}a), we analysed final-layer token embeddings produced by Qwen2.5–7B and its core-lesioned counterpart. For each language (English, Chinese, French), token-level embeddings were extracted from model encoding outputs together with token metadata.

Subword tokens were aggregated into word-level representations by grouping tokens sharing the same \texttt{word\_idx} and applying mean pooling across their embeddings \cite{bommasani2020interpreting}.

To characterise global representational structure while ensuring computational tractability, we applied a two-stage dimensionality reduction pipeline. Principal Component Analysis (PCA) \cite{abdi2010principal} first projected embeddings to 50 dimensions, followed by Uniform Manifold Approximation and Projection (UMAP) \cite{mcinnes2018umap} to obtain two-dimensional representations. This procedure was applied independently to intact and lesioned models to enable direct comparison of their embedding geometries.

\paragraph{Multilingual probing evaluation}

We evaluated linguistic information encoded in sentence representations using the SentEval probing framework shown in Figure \ref{fig:core-language-ablation}c. Four structurally grounded tasks were examined across English, French, and Chinese: sentence length, syntactic tree depth, top constituents, and coordination inversion. All datasets followed the standard SentEval \cite{conneau2018senteval} format with predefined train, development, and test splits.
Multilingual counterparts were constructed by translating the original English probing datasets into French and Chinese. 

Because several probing labels depend on surface or syntactic properties, labels were recomputed on the translated corpora. Sentence length labels were reassigned based on token counts in the target language, with segmentation adapted to language-specific tokenisation conventions.

Sentence embeddings were extracted from hidden states and mean-pooled across non-padding tokens using the attention mask \cite{bommasani2020interpreting}. 
Linguistic information was assessed by training supervised classifiers on frozen sentence embeddings. We used a two-layer multilayer perceptron (Linear → ReLU → Dropout → Linear; hidden size 128). Models were trained with Adam \cite{kingma2014adam} (learning rate $1\times10^{-3}$) for 20 epochs, with model selection based on development-set accuracy. Final performance was reported on held-out test data using accuracy and macro-averaged precision, recall, and F1 metrics.

\subsection*{Neural Encoding Analysis} 

Neural encoding analyses were conducted via a pipeline linking large language model (LLM) representations to fMRI responses \cite{schrimpf2021neural}. First, contextualized token embeddings were extracted from the final hidden layer of each intact or lesioned LLM for the time-stamped narrative transcripts. Second, token-level embeddings were temporally aligned to the fMRI acquisition by averaging all embeddings occurring within each repetition time (TR) \cite{huth2016natural}. Third, to account for the hemodynamic response delay, the resulting embedding time series were shifted by 4 seconds relative to the BOLD signal \cite{jain2018incorporating}. Fourth, voxel-wise ridge regression models were trained independently for each participant to predict BOLD time series from the lagged embeddings using run-wise cross-validation. Finally, for each participant and each voxel, a prediction accuracy was quantified as the Pearson correlation between predicted and observed BOLD responses \cite{schrimpf2021neural, kay2008identifying}.

\paragraph{fMRI Dataset and Preprocessing} 
We used the Le Petit Prince multilingual naturalistic fMRI corpus (LPPC-fMRI), an open-access dataset designed to study neural mechanisms of language comprehension under naturalistic listening conditions \cite{li2022petit}. The corpus includes fMRI recordings from 112 healthy, right-handed participants (49 native English, 35 native Chinese, and 28 native French speakers), each listening to an audiobook version of The Little Prince in their native language while undergoing multi-echo fMRI scanning. The total listening duration was approximately 100 minutes, divided into nine runs of about ten minutes each.

Functional MRI data were acquired using 3T MRI scanners. English and Chinese data were collected on a GE Discovery MR750 system (32-channel coil), while French data were collected on a Siemens Magnetom Prisma Fit system (64-channel coil). Functional data were obtained with a multi-echo EPI sequence (TR = 2000 ms; TE = 2.8, 27.5, 43 ms for English and Chinese; TE = 10, 25, 38 ms for French; flip angle = 77°; voxel size = 3.75 × 3.75 × 3.8 mm³).

Preprocessing was performed using the AFNI (v16) \cite{cox1996afni} and ME-ICA \cite{kundu2012differentiating, kundu2013integrated} pipelines. The steps included slice-timing correction, despiking, motion correction, nonlinear registration to the MNI template, and denoising via multi-echo independent component analysis to remove motion, physiological, and scanner artifacts. Data were resampled to 2 mm isotropic voxels.
In our preprocessing, the BOLD time series for each voxel were z-score normalized \cite{pereira2009machine}.

To restrict analyses to neuroanatomically relevant regions and to reduce computational burden, we applied a standardized cortical gray-matter mask derived from the MNI152 Template space \cite{woolrich2009bayesian,smith2004advances}. Specifically, we used the MNI152\_template\_gm\_mask\_2mm, which delineates voxels with high probability of belonging to cortical gray matter at 2 mm isotropic resolution \cite{mazziotta2001probabilistic,fonov2011unbiased}.
The mask was applied after spatial normalization and resampling, ensuring voxel-wise correspondence across participants in MNI space. By excluding non-gray-matter voxels—such as white matter, cerebrospinal fluid (CSF), and subcortical structures—we limited the encoding analysis to cortical tissue most strongly implicated in high-level language processing and narrative comprehension \cite{zhang2002segmentation}.

\paragraph{Extracting LLM Representations} The time-stamped transcripts served as input to all intact and ablated LLMs. We extracted contextualized embeddings for each token from the final hidden layer of each model. The context for each token consisted of all preceding complete sentences and the current sentence up to the 512th token to its left, consistent with the causal attention mechanism of decoder-only models. To match the temporal resolution of the fMRI data, we averaged the embeddings of all tokens occurring within each scan's time interval (TR) \cite{huth2016natural}. To account for the hemodynamic lag, these aggregated embedding time series were delayed by 4 seconds relative to the BOLD signal time series \cite{jain2018incorporating}. 
\paragraph{Voxel-wise Encoding Model} For each voxel in the brain, we independently trained a ridge regression model to predict its BOLD signal time series from the corresponding lagged LLM embedding sequence \cite{naselaris2011encoding}, with the regularization parameter chosen to prevent overfitting. We employed a 9-fold cross-validation scheme, where the data from the 9 acquisition runs were used iteratively as training (8 runs) and test (1 run) sets. This process yielded a full predicted BOLD time series for each voxel. The final encoding performance for each participant at each voxel was computed by first calculating the correlation between the predicted and actual BOLD signals for each of the 9 folds and then averaging these 9 correlation values. 

\subsection*{Evaluation and Statistical Analysis} 

\paragraph{Performance Metric and Statistical Comparison} We used the Pearson correlation coefficient ($r$) to quantify the encoding performance, measuring the correspondence between the model-predicted and the observed BOLD signals in the test sets. To assess the causal impact of the lesions, we performed a paired t-test at each voxel comparing the Fisher-transformed correlation coefficients obtained from the intact model ($r_{\text{intact}}$) and each ablated model ($r_{\text{ablated}}$) \cite{friston1994statistical}. A significant positive t-statistic indicates that ablating a specific subnetwork significantly impairs the model's ability to predict neural activity in that voxel, suggesting a crucial functional contribution of that subnetwork. 
\paragraph{Multiple Comparisons Correction and Visualization} To control the false positive rate across the tens of thousands of voxels tested, we applied a significance threshold of $\mathrm{FDR}<0.01$. For visualization, statistically significant results were projected as heatmaps onto the MNI152 standard brain template \cite{mazziotta1995probabilistic, mazziotta2001probabilistic, mazziotta2001four}. The analysis was restricted to cortical voxels using the MNI152\_template\_gm\_mask\_2mm grey matter mask.

\section{Data availability}

The neuroimaging data used in this study were obtained from the publicly available \textit{Le Petit Prince} multilingual fMRI corpus, which provides naturalistic story-listening data across multiple languages using ecologically valid stimuli \cite{li2022petit}. The dataset is accessible via OpenNeuro at: \url{https://openneuro.org/datasets/ds003643/versions/2.0.0}.

Cortical surface masks were derived from the ICBM152 nonlinear 2009 template provided by the Montreal Neurological Institute (MNI), available at: \url{https://www.bic.mni.mcgill.ca/ServicesAtlases/ICBM152NLin2009}.

Language-selective cortical parcels were obtained from the functional localization resources released by the MIT EvLab \cite{malik2022investigation}, available at: \url{https://www.evlab.mit.edu/resources-all/download-parcels}.

All datasets are publicly available for research use under their respective data usage agreements.

\section{Code availability}
All code supporting this study is publicly available at:
\url{https://github.com/yang-C23/Shared-Unique-Neural-Architecture-for-Language}.
The repository contains scripts for multilingual embedding extraction, structured parameter ablation, neural encoding model training, statistical analysis, and figure generation. Documentation and environment configuration files are provided to facilitate reproducibility.

\section*{Declarations}
\paragraph{Acknowledgements}
This work was supported by computational resources provided through the EuroHPC Joint Undertaking. The authors acknowledge access to the Leonardo supercomputing system, hosted by CINECA, through the EuroHPC project Aligning Brain Language Representation with Large Language Models: Benchmark and Scalability Test (Account ID: B24\_036; PI: Goran Nenadic; project code: EHPC-BEN-2025B05-036). These resources were essential for large-scale model inference, computational lesion analyses, and neural encoding experiments. We also thank CINECA for infrastructure provision and technical support. Yang Cui was supported by a PhD fee-waiver studentship from the Department of Computer Science, Faculty of Science and Engineering, The University of Manchester, for the PhD programme in Computer Science.

\paragraph{Author contributions}

Y.C. designed the study, developed the computational framework and ablation algorithms, performed the analyses, generated visualizations, and wrote the manuscript. J.S. conceived and designed the study, supervised the research, and co-wrote the manuscript. Y.S. contributed to results visualization and figure preparation. Y.W. contributed to manuscript revision and editing. Y.Z. contributed to mask processing, and manuscript revision. J.L. contributed to experimental design, data processing, and manuscript revision. S.W. contributed to experimental design and manuscript revision. J.H. contributed to experimental design and manuscript revision. G.N. supervised the project, contributed to experimental design, and manuscript revision.



\noindent






\begin{appendices}

\section{}
\renewcommand{\thefigure}{A\arabic{figure}}
\setcounter{figure}{0}

\begin{figure*}[t]
\centering
\includegraphics[width=\textwidth]{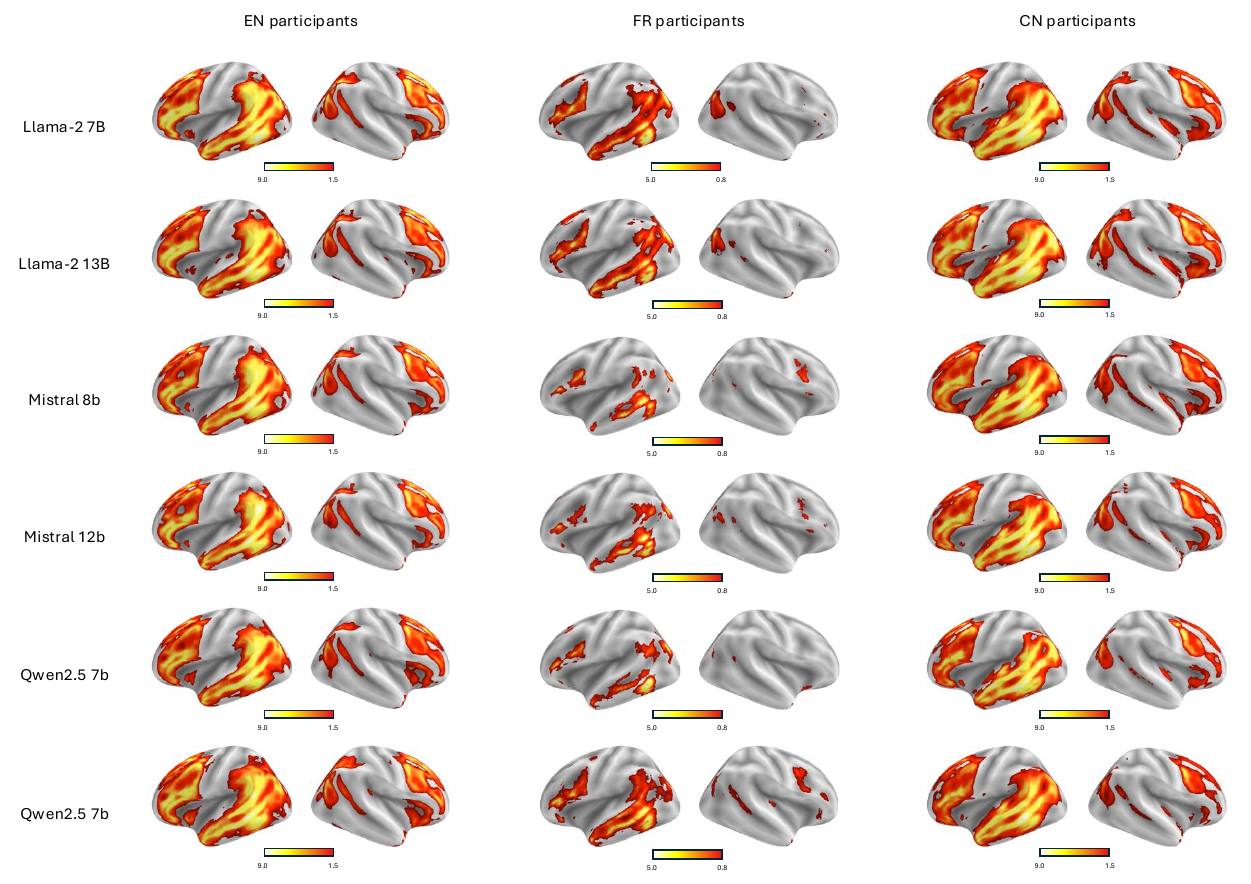}
\caption{
\footnotesize
\textbf{Cortical encoding t-maps comparing intact and core-language-ablated models across multilingual participants.}
Voxel-wise t-maps from six large language models are shown for intact and core-language-ablated conditions across English-, Chinese-, and French-native participant groups. Intact models exhibit robust encoding throughout the distributed language network, whereas core ablation produces widespread attenuation of encoding strength, particularly in superior temporal and inferior frontal cortices. Lesion effects are spatially convergent across participant groups. Maps are thresholded at $\mathrm{FDR} < 0.01$ (voxel-wise, cortical mask).
}
\label{fig:appendix_tmap_comparison}
\end{figure*}

\end{appendices}






\clearpage


\end{document}